# Feedback U-net for Cell Image Segmentation

Eisuke Shibuya
Meijo University
1-501 Shiogamaguchi, Tempaku-ku, Nagoya
468-8502, Japan
160442066@ccalumni.meijo-u.ac.jp

Kazuhiro Hotta
Meijo University
1-501 Shiogamaguchi, Tempaku-ku, Nagoya
468-8502, Japan
kazuhotta@meijo-u.ac.jp

## Abstract

*Human brain is a layered structure, and performs not only a feedforward process from a lower layer to an upper layer but also a feedback process from an upper layer to a lower layer. The layer is a collection of neurons, and neural network is a mathematical model of the function of neurons. Although neural network imitates the human brain, everyone uses only feedforward process from the lower layer to the upper layer, and feedback process from the upper layer to the lower layer is not used. Therefore, in this paper, we propose Feedback U-Net using Convolutional LSTM which is the segmentation method using Convolutional LSTM and feedback process. The output of U-net gave feedback to the input, and the second round is performed. By using Convolutional LSTM, the features in the second round are extracted based on the features acquired in the first round. On both of the Drosophila cell image and Mouse cell image datasets, our method outperformed conventional U-Net which uses only feedforward process.*

## 1. Introduction

Human brain is known to have a layered structure, and the contents of layer are the collection of nerve cells called neurons. In addition to feedforward processing from the lower layer handling low-level information to the upper layer handling high-level information, feedback processing from the upper layer to the lower layer is also performed. Neurons are good at information processing and propagation. A neuron processes the information received from a large number of adjacent neurons. When the result exceeds a threshold value, the neuron transmits the result to next neuron. If the result does not exceed the threshold value, it is determined that it is not important information and the information is not propagated. A mathematical model of a neuron [1] is called a neural network, and a complex function approximation is possible by connecting many layers. Neural network updates the weight of each layer so that the difference from labels becomes small. In such neural network, since each neuron is fully connected, positional information such as an image are lost. Therefore, a convolutional neural network [2] with convolutional layers and pooling layers is effective for image recognition.

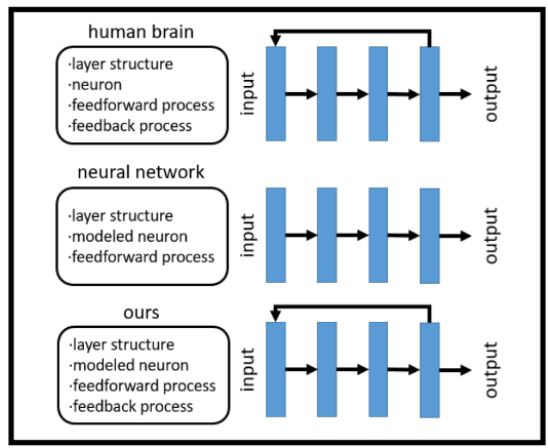

Fig. 1: Top row shows the structure of human brain. Middle row shows the structure of neural network, and bottom row shows the structure of our method.

Recently, the development of CNN has been successful in various tasks such as image classification [3], semantic segmentation [4], object detection [5] and object tracking [6], and image generation [7]. Convolutional layer makes it possible to acquire features while maintaining spatial information. Pooling layer compresses information and performs downsampling to obtain position invariance. By repeating these two layers, high-level features can be extracted, and the accuracy is improved. However, the increase in the number of layers causes the vanishing gradient problem and the degradation problem. This problem has been solved by ResNet [8]. Since then, many researchers have been focusing on deepening the network for better performance. In addition, attention mechanisms [9] that focus on important parts in feature maps can also be used for better performance. Squeeze-and-Excitation Networks [10], a kind of attention mechanism, is very useful because it can be used in various models.

In recent years, various modes have been proposed for CNN that imitates the human brain, but feedback processing from the upper layer to the lower layer is not used well for better performance. Since feedback is used in the visual cortex, we consider that the accuracy will be improved by incorporating it into CNN. In this paper, we proposed Feedback U-Net using Convolutional LSTM.

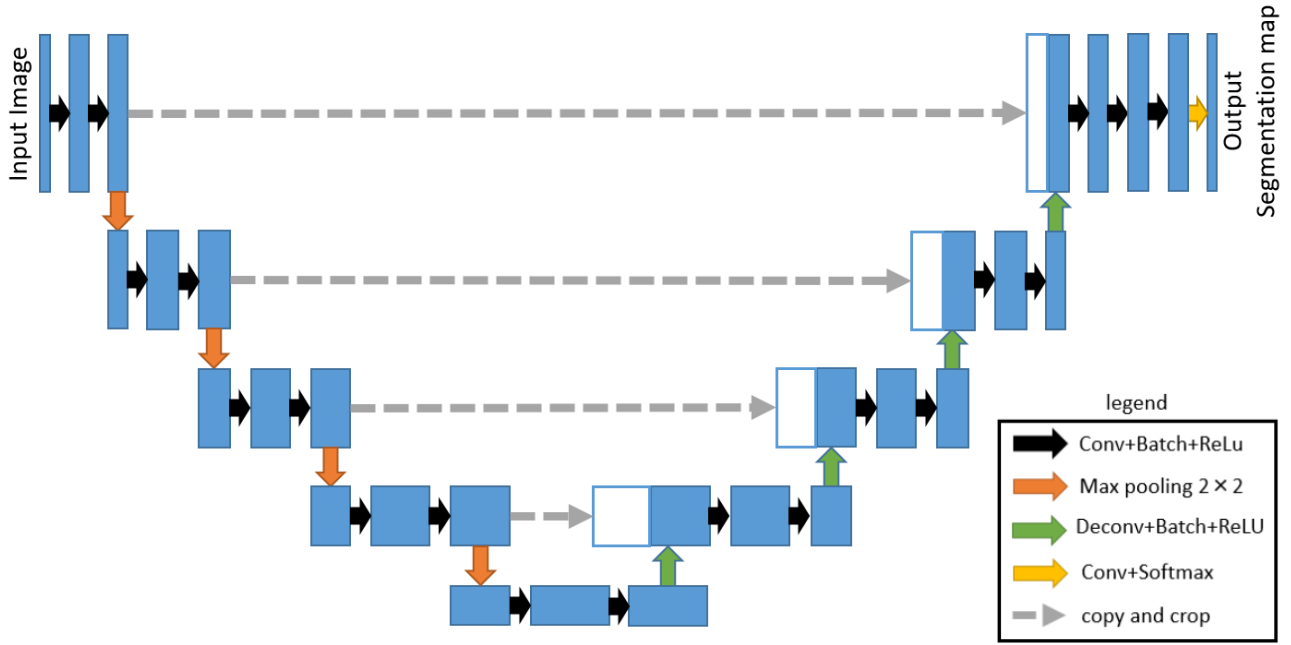

Fig. 2: U-Net architecture

Bottom row in Fig. 1 shows our method. Our approach is the only one method which feeds back the output obtained once to the input layer of the network again. Since the same layers are used twice, we use convolutional LSTM [11] which deals with sequential data. We maintain the features extracted in the first round, and extract features in the second round based on the features in the first round. Our proposed Feedback U-net outperformed conventional U-Net which uses only feedforward process.

This paper is organized as follows. Section 2 describes the related works. The architecture of the proposed Feedback U-Net using Convolutional LSTM is presented in section 3. Section 4 shows the experimental results on two kinds of cell image datasets. Finally, conclusion and future works are described in section 5.

## 2. Related Work

### 2.1. Semantic segmentation

Semantic segmentation is a task for assigning class labels to each pixel in an image. Segmentation is used in various fields such as in-vehicle cameras and medical image processing. The recent semantic segmentation methods using deep learning are based on fully convolutional network (FCN) [12]. FCN did not require fully connected layers, and allows segmentation on images of any size.

Encoder-decoder structure is also used in semantic segmentation. It composed of encoder network that extracts features using convolutional layers and pooling layers, and decoder network that performs classification based on the extracted features. Encoder extracts features from the input image by convolution and pooling layers, and finally obtains global features with low resolution. Decoder restores the global features obtained by the encoder to the original image size using convolution and upsampling layers. The network is devised to supplement the location information lost by pooling layers. SegNet [13] copies the indices selected by max pooling to the decoder. This not only allows the decoder to complement the upsampling location information, but also makes the memory more efficient than copying the feature maps.

Another famous model is the U-Net [14]. U-Net was proposed for medical image segmentation, and one of the most famous CNN models. Fig. 2 shows the structure of the U-Net used in this paper. At convolutional layers, we used ReLU activation function in common with encoder and decoder. In the encoder, max pooling is used for downsmpling. In the decoder, deconvolution is used for upsampling. The most important characteristic of U-Net is skip connection between encoder and decoder. The feature map with the position information in the encoder is concatenated to the restored feature map in the decoder. Therefore, the position information is complemented, and each pixel can be more accurately assigned to the class label. In addition, an improved model has been proposed. U-

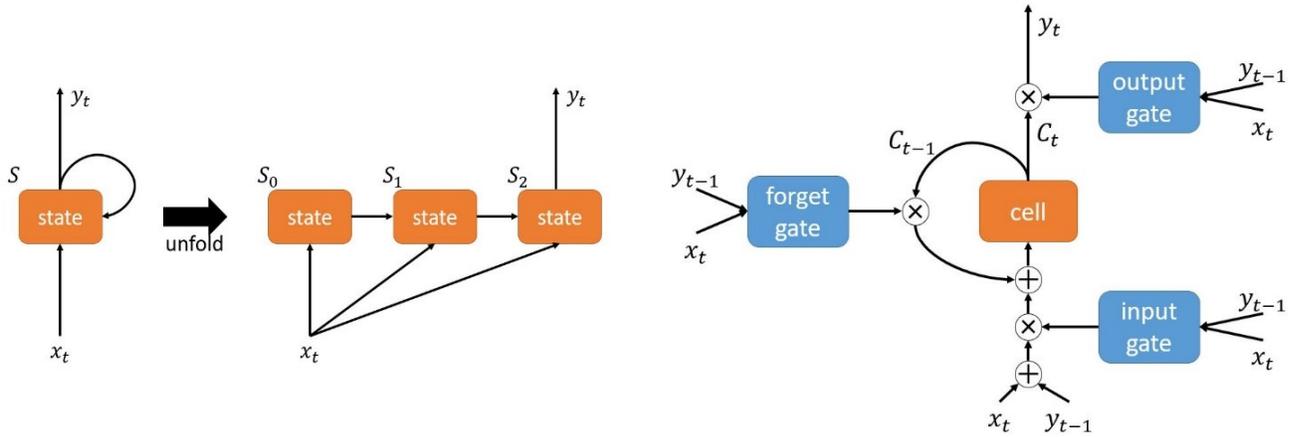

Fig. 3: Recurrent convolutional layers and convolutional LSTM. Left shows recurrent convolutional layer. Right shows convolutional LSTM which consists of input gate, forget gate, output gate, and cell.

Net++ [15] integrates multi-scale features. Attention U-net [16] used attention in skip connection. U-net is effective for segmentation but U-Net based methods used only feedforward processing from the lower layer to the upper layer. In this paper, we add feedback processing to U-net newly.

## 2.2. Conventional methods using feedback

There is no model that feeds back the output of the network to input, but there are several approaches to feed back layer's output. RU-Net [17] is a medical image segmentation model composed of U-Net and recurrent neural network. RU-Net replaces each convolutional layer with recurrent convolutional layer [18]. Recurrent convolutional layer is a model that the concept of recurrent neural network is adapted to convolutional layer. Fig. 3 left shows recurrent convolutional layer. In recurrent convolutional layer, the value of state is fed back, and the value is added to the next state. RU-Net repeatedly performs convolution at each scale in recurrent convolutional layer and accumulates feature information. Therefore, feature representation is better than standard convolution. However, since RU-Net repeatedly performs convolution with the same input as shown Fig. 3, we see that it is not feedback but deepening of network. Furthermore, even if the output of network is fed back in this model, convolution of the first and second rounds is performed independently.

Our approach uses convolutional LSTM instead of recurrent convolutional layer. Convolutional LSTM is convolutional version of LSTM [19], and it deals sequential data. Convolutional LSTM consists of input gate, output gate, forget gate, and cell as shown in Fig. 3. By adding the gate that controls input and output to the conventional recurrent neural network, long-term dependent has been solved. Especially, forget gate [20] has the ability to forget unnecessary information from the features maintained in the cell. Convolutional LSTM is used for predicting the movement of rain clouds [21].

In this paper, the sequential information of the first and second rounds is used. The features extracted in the first round are maintained in the cell, and the features in the second round are extracted based on the maintained features.

## 3. Feedback U-Net with convolutional LSTM

### 3.1. Architecture

Fig. 4. illustrates the proposed method. We made two major changes to U-Net. The first change is to do feedback the output of U-net to input layer. The second change is the usage of convolutional LSTM. The details are explained as follows. In the U-Net, we acquire probability map of each class by a softmax function at the final layer. In our model, the probability map of each class is fed back to input layer.

For example, in the case of segmentation of 4 classes, 4 probability maps are fed back to input layer. Thus, the input of U-net for the second round is the segmentation result at the first round. The final segmentation result is obtained as the output of U-net at the second round. Note that we use the same convolutional layers for both rounds. However, we use different batch normalization for each round as shown in Fig. 5.

Our model incorporates 6 convolutional LSTM. Since convolutional LSTM has the function that maintains features extracted before, it is possible to perform convolution based on the features extracted at the first round. When feedback is performed in normal convolutional layer that does not deal with sequential data, only weights are shared. Thus, the features extracted at the first round is unrelated to the features extracted at the second round. In contrast, our approach replaces convolutional layer with convolutional LSTM. When we extract features at the second round, the features at the first round are also used so that more useful features can be obtained. In this paper, we put convolutional LSTM at the locations where local and global features are available. Fig. 4 a, b, c, d and e shows the locations. It is common for two kinds of cell image datasets used in experiments. In location a, b, d, and e, resolution is the highest and they have local features with position information. Thus, the locations a, b, d, and e attempts to complement classes with small area.

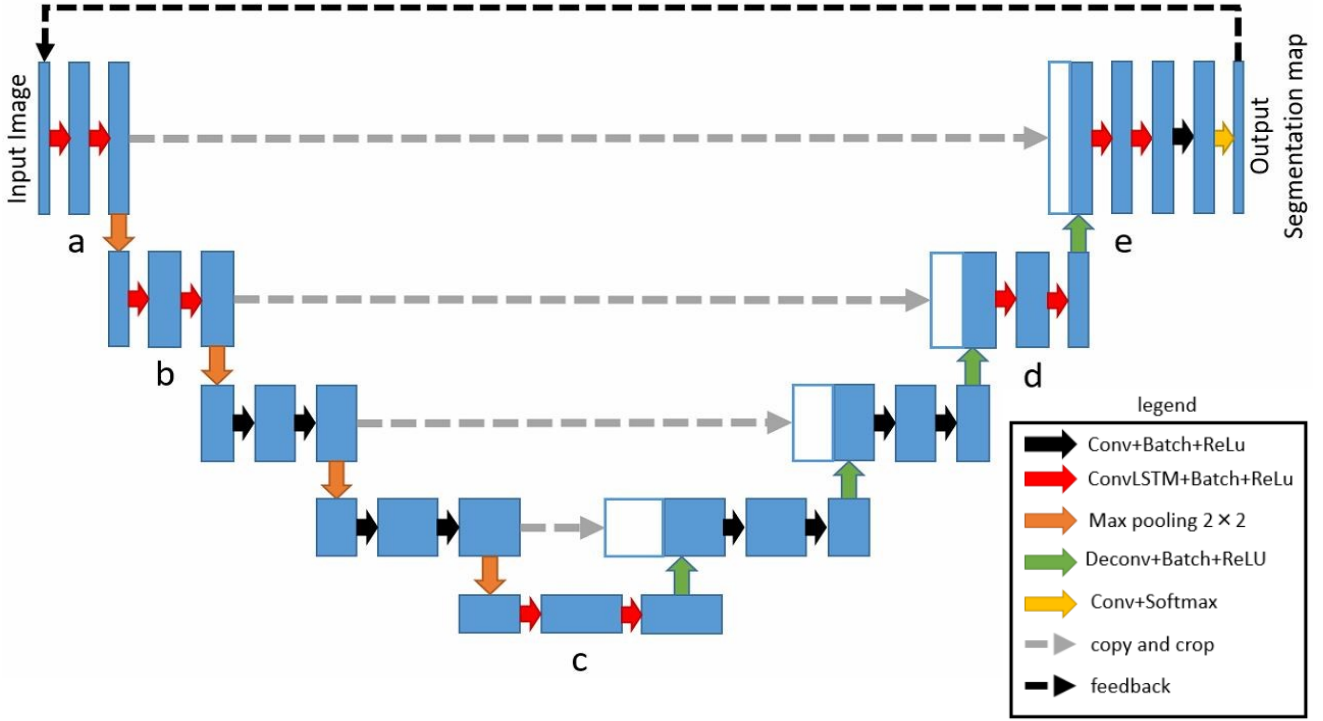

Fig. 4: Feedback U-Net with Convolutional LSTM

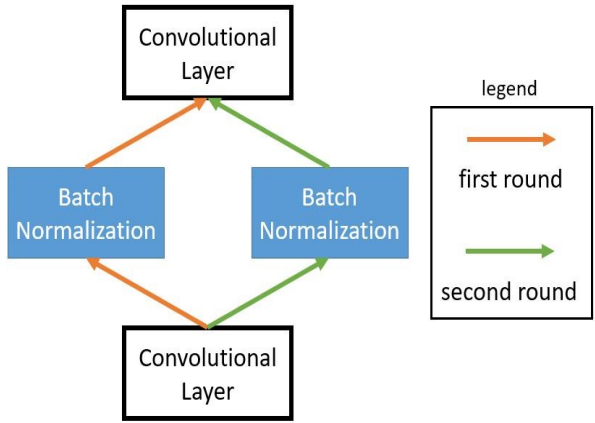

Fig. 5: The same convolutional LSTM layer is used for both rounds. We used different batch normalization for each round.

### 3.2. Loss function

Our model is trained with 2 loss functions; the loss for the first round ($L_{first}$) and the second round ($L_{second}$). Both of them are defined as softmax cross entropy loss.

$$L_{first} = L_{second} = -\sum_i \sum_c p_c^i \log q_c^i, \quad (1)$$

where $i$ means the i-th sample in dataset, $c$ means the c-th class, $p_c^i$ is one hot vector of ground truth, $q_c^i$ is the probability of class $c$ for the i-th sample. The overall loss is given by

$$L = \lambda L_{first} + L_{second} \quad (2)$$

where $\lambda$ is a hyperparameter. In this paper, we set $\lambda$ to 0.5 because the second round is more important for segmentation.

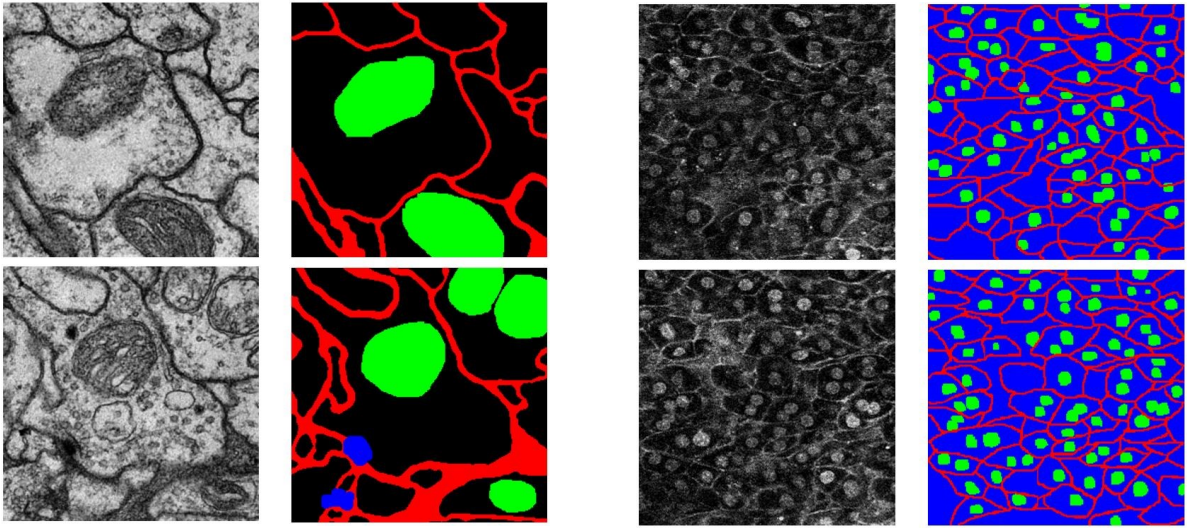

Fig. 6: Examples of datasets. Left shows Drosophila cell image dataset which consists of cytoplasm, cell membrane, mitochondria, and synapses. Right shows mouse cell image dataset which consists of cytoplasm, cell membrane, and cell nucleus.

## 4. Experiments

### 4.1. Datasets and metrics

We use Drosophila cell image dataset [22] as shown in left two columns of Fig. 6. The dataset consists of 4 classes; cytoplasm, cell membrane, mitochondria and synapses. Since the original size is 1024×1024 pixels, we cropped a region of 256×256 pixels from original images due to the size of GPU memory. There is no overlap for cropping areas, and the total number of crops is 320. We used 192 regions for training, 48 for validation and 80 for test. We evaluate our method with 5 fold cross-validation.

We also use mouse cell image dataset as shown in two right columns of Fig. 6. The dataset consists of 3 classes; cytoplasm, cell membrane and cell nucleus. We did data augmentation which includes 90 degrees rotations and left-right flip. By the augmentation, we have 400 images from 50 original images. We used 280 for training, 40 for validation and 80 for test. We evaluate 8 fold cross-validation.

In semantic segmentation, Intersection over Union (IoU) is used as evaluation measure. IoU is the overlap ratio between prediction and ground truth labels. In this paper, we use IoU of each class and mean IoU which is the average IoU of all classes.

### 4.2. Implementation details

In this paper, we use keras library and train network using Adam for 1500 epochs with a learning rate of 0.0001. Batch size is set to 16 for the Drosophila cell image dataset, and 10 for the Mouse cell image dataset. Furthermore, class weight is used to solve class imbalance problem. The number of filters at convolution and convolutional LSTM layers is set to 8, 16, 32, 64 and 128 from the top to bottom of the U-net.

We compare 5 methods; conventional U-Net, RU-Net, Feedback U-Net with recurrent convolutional layer, Feedback U-Net without convolutional LSTM, the proposed Feedback U-Net with convolutional LSTM. RU-Net is only the conventional method using recurrent convolutional layer and U-Net. Time-step for RU-net is set to 2. This is the same with original paper [18]. Feedback U-Net with recurrent convolutional layer is the model which replaces convolutional LSTM in our approach with recurrent convolutional layer to show the effectiveness of convolutional LSTM.

### 4.3. Comparison with Another Method

In Table 1, we compare U-Net with our proposed Feedback U-Net with/without convolutional LSTM on Drosophila cell image dataset. Our method achieved the best accuracy which is 71.5% on mean IoU. RU-Net which is conventional method using recurrent convolution and U-net provided 71.4%. In contrast, for Feedback U-Net with recurrent convolutional layer and Feedback U-Net without Convolutional LSTM, there is no significant improvement in accuracy over the baseline. Especially, we confirmed that the accuracy of synapses with small area is reduced. We consider that high-level features are obtained regardless of the presence or absence of convolutional LSTM by performing feedback processing for 3 classes with large area such as cytoplasm, cell membrane and mitochondria. Thus, IoU of those classes increased. However, for synapse class with small area, high-level features are lost without convolutional LSTM and IoU decreased.

Table 1: Comparison result on the Drosophila cell image dataset.

| Method | cytoplasm (%) | cell membrane (%) | mitochondria (%) | synapses (%) | meanIoU (%) |
|---|---|---|---|---|---|
| U-Net | 92.0 | 74.9 | 73.1 | 40.9 | 70.2 |
| RU-Net (time-step=2) | 92.2 | 75.8 | 74.3 | **43.2** | 71.4 |
| Feedback U-Net with Recurrent Neural Layer | 92.2 | 75.1 | 73.8 | 37.9 | 69.8 |
| Feedback U-Net without Convolutional LSTM | 92.1 | 76.0 | 74.9 | 38.4 | 70.4 |
| Feedback U-Net with Convolutional LSTM | **92.4** | **76.4** | **75.2** | 42.3 | **71.5** |

Table 2: Comparison result on the Mouse image dataset.

| Method | cytoplasm (%) | cell membrane (%) | cell nucleus (%) | meanIoU (%) |
|---|---|---|---|---|
| U-Net | 77.1 | 33.8 | 65.0 | 58.6 |
| RU-Net (time-step=2) | 77.0 | 33.8 | 63.7 | 57.7 |
| Feedback U-Net with Recurrent Neural Layer | 77.1 | 34.4 | 64.1 | 58.6 |
| Feedback U-Net without Convolutional LSTM | 77.0 | 33.8 | 63.7 | 58.1 |
| Feedback U-Net with Convolutional LSTM | **77.1** | **35.9** | **65.6** | **59.5** |

In Table 2, we also evaluate our method on mouse cell image dataset. The proposed method achieved the best accuracy 59.5% on mean IoU. Other methods do not improve the accuracy from U-Net. In addition, our approach has higher generalization ability than RU-Net. From the results on two kinds of cell image datasets, the effectiveness of our method is demonstrated.

### 4.4. Qualitative Results

Fig. 7 shows the segmentation results by each method. From left to right, input image, ground true image, the results by U-Net, Feedback U-Net without convolutional LSTM, and Feedback U-Net with convolutional LSTM are shown. In the case of the Drosophila cell image dataset, Feedback U-Net without convolutional LSTM is better than U-Net on distinction between cell membrane and mitochondria. However, undetected area of synapses stands out. In contrast, our approach gave good segmentation result for all classes. The cell membrane and mitochondria are well distinguished, and there is little false detection of synapses. In the case of the Mouse cell image dataset, there is no noticeable difference between U-Net and Feedback U-Net without convolutional LSTM, and cell membrane is severely broken. However, our approach improves the accuracy of cell membrane and detects more connected membrane.

Fig. 8 shows the sum of the outputs of the first convolutional layer or convolutional LSTM layer on the second round. ReLU function is used after convolution. From left to right shows that ground truth image, the output of Feedback U-Net without convolutional LSTM, and the output of our method. It turns out that our approach can extract the feature map highlighted with cell membrane, cell nucleus, mitochondria, and synapses. In contrast, the feature map of Feedback U-Net without convolutional LSTM losses the information of cell membrane, mitochondria, and especially synapses. According to these results, we consider that our approach complements for the features of object class not background in the second round.

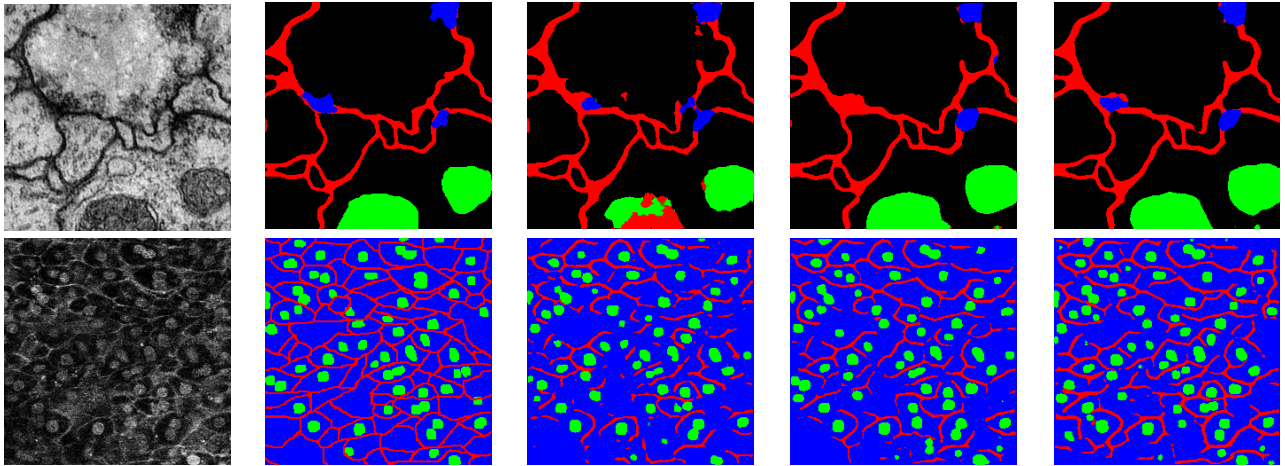

Fig. 7: Qualitative Results. From left to right, input image, ground truth image, the result by U-Net, Feedback U-Net without convolutional LSTM and our proposed method.

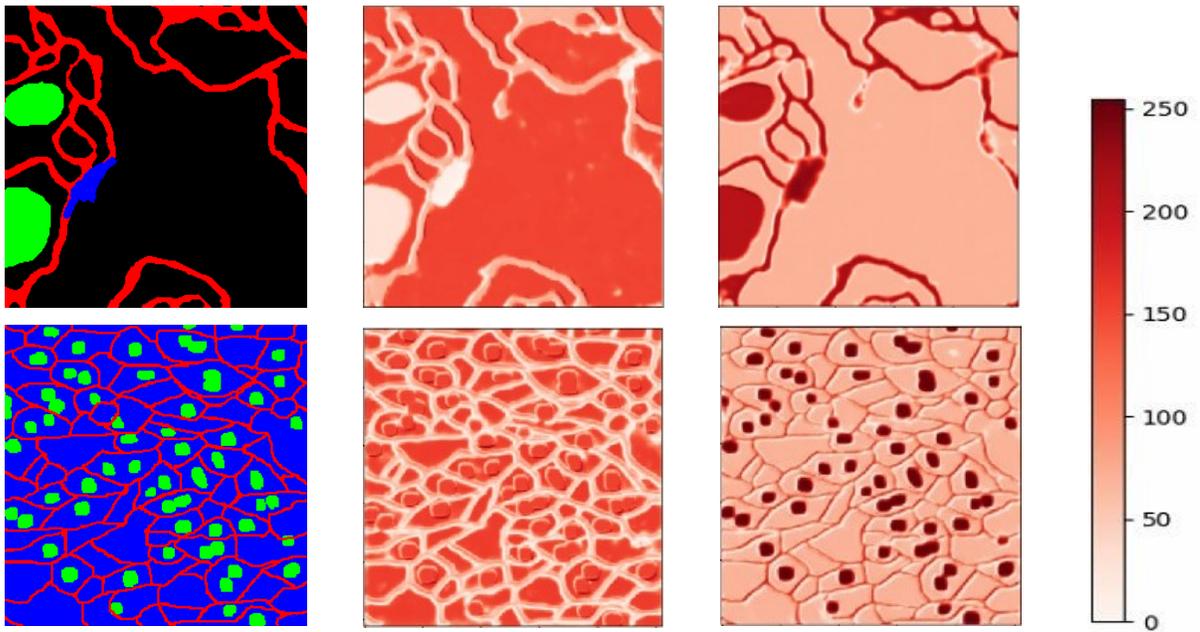

Fig. 8: The sum of outputs of the first convolutional layer or convolutional LSTM layer on the second round. From left to right shows that ground truth, Feedback U-Net without convolutional LSTM and our method.

This is because our proposed method outperformed conventional methods.

### 4.5. Ablation Study

In Table 3 and 4, we conduct an ablation study about the locations of convolutional LSTM. Note that locations are shown in Fig. 4. When we implement convolutional LSTM in only the location of Fig. 4 (e), the most global information is available. By comparing with ablation c, it can be determined whether local information or global information should be maintained. Below the dashed line, the accuracy of our full model is shown.

We see that the position of (a) is the most important. In, addition, the position of (e) is the secondary important. These two are positions where image size is the biggest, and they have local feature information with correct position. Therefore, it is possible to extract small features like synapses. Actually, the accuracy of synapses is influenced by convolutional LSTM at position (a) and (e). It turns out that it is better to maintain local features than global features by comparing with (c) and (a, b, d, and e).

Table 3: Ablation study on the Drosophila cell image dataset.

| Ablation | cytoplasm (%) | cell membrane (%) | mitochondria (%) | synapses (%) | meanIoU (%) |
|---|---|---|---|---|---|
| a | 92.2 | 75.7 | 73.5 | 37.4 | 69.7 |
| b | 92.2 | 75.3 | 73.6 | 42.5 | 70.9 |
| c | 92.1 | 75.6 | 74.6 | 40.8 | 70.8 |
| d | 92.4 | 75.9 | 74.6 | 42.1 | 71.3 |
| e | 92.4 | 76.1 | 74.6 | 38.0 | 70.3 |
| a, b, d, e | 92.2 | 75.8 | 73.1 | 33.9 | 68.6 |
| ours | **92.4** | **76.4** | **75.2** | **42.3** | **71.5** |

Table 4: Ablation study on the Mouse cell image dataset.

| Ablation | cytoplasm (%) | cell membrane (%) | cell nucleus (%) | meanIoU (%) |
|---|---|---|---|---|
| a | 77.0 | 35.1 | 64.6 | 58.9 |
| b | 77.0 | 35.5 | 64.5 | 59.0 |
| c | 76.9 | **36.6** | 64.4 | 59.3 |
| d | 76.9 | 35.3 | 64.5 | 58.9 |
| e | 77.1 | 34.8 | 65.1 | 59.0 |
| a, b, d, e | 77.0 | 33.7 | 64.1 | 58.2 |
| ours | **77.1** | 35.9 | **65.6** | **59.5** |

## 5. Conclusion

In this paper, we proposed Feedback U-Net with convolutional LSTM which used feedback process like human brain. Our results demonstrated that the combination of feedback process from output layer to input layer and convolutional LSTM layer which handles sequential data is a valid to segmentation. Convolutional LSTM makes it possible to extract feature map of object classes (e.g. cell membrane, cell nucleus, mitochondria, and synapses) not background. Especially, classes with small area are influenced by position where convolutional LSTM is used. There may be better placement pattern of convolutional LSTM than our approach. This is a subject for future works.

**Acknowledgments**

This work is partially supported by MEXT/JSPS KAKENHI Grand Number 18H04746 "Resonance Bio" and 18K11382.